\documentclass[letterpaper, 10 pt, conference, hidelinks]{ieeeconf}  

\IEEEoverridecommandlockouts                              

\usepackage{mathptmx} 
\usepackage{times} 
\usepackage{amsmath} 
\usepackage{amssymb}  
\usepackage{mathtools}
\usepackage{balance}
\usepackage{graphicx}
\usepackage{nth}
\usepackage{siunitx} 
\usepackage{bm} 
\usepackage{subcaption} 
\usepackage[table]{xcolor} 
\usepackage{tabularx} 
\usepackage{dcolumn} 
\usepackage{algorithm}
\usepackage[noend]{algpseudocode}

\DeclareMathOperator*{\argmin}{arg\,min}

\title{\LARGE \bf
Inverse Statics Optimization for Compound Tensegrity Robots
}

\author{Andrew P. Sabelhaus$^{1}$, Albert H. Li$^{1}$, Kimberly A. Sover$^{1}$, \\ Jacob Madden$^{1}$, Andrew Barkan$^{1}$, Adrian K. Agogino$^{2}$, Alice M. Agogino$^{1}$
\thanks{$^{1}$A.P. Sabelhaus, A.H. Li, K.A. Sover, J. Madden, A. Barkan, and A.M. Agogino are with the Department of Mechanical Engineering, University of California Berkeley, USA.
        {\tt\small \{apsabelhaus, alberthli, kasover, jmadden, andrew\_barkan, agogino\} @berkeley.edu}}%
\thanks{$^{2}$A.K. Agogino is with the Intelligent Systems Division, NASA Ames Research Center, Moffet Field CA 94035.
        {\tt\small adrian.k.agogino@nasa.gov }}%
}

\begin{document}


\newcommand{\bzero}{\mathbf{0}} 
\newcommand{\bOnes}{\mathbf{1}} 
\newcommand{\bA}{\mathbf{A}} 
\newcommand{\ba}{\mathbf{a}} 
\newcommand{\bB}{\mathbf{B}} 
\newcommand{\bb}{\mathbf{b}} 
\newcommand{\bC}{\mathbf{C}} 
\newcommand{\bc}{\mathbf{c}} 
\newcommand{\bD}{\mathbf{D}} 
\newcommand{\bd}{\mathbf{d}} 
\newcommand{\bE}{\mathbf{E}} 
\newcommand{\be}{\mathbf{e}} 
\newcommand{\bF}{\mathbf{F}} 
\newcommand{\btF}{\mathbf{\widetilde F}} 
\newcommand{\boldf}{\mathbf{f}} 
\newcommand{\boldtf}{\mathbf{\widetilde f}} 
\newcommand{\bG}{\mathbf{G}} 
\newcommand{\btG}{\mathbf{\widetilde G}} 
\newcommand{\bg}{\mathbf{g}} 
\newcommand{\bH}{\mathbf{H}} 
\newcommand{\bh}{\mathbf{h}} 
\newcommand{\bI}{\mathbf{I}} 
\newcommand{\bJ}{\mathbf{J}} 
\newcommand{\bK}{\mathbf{K}} 
\newcommand{\bL}{\mathbf{L}} 
\newcommand{\bell}{\bm{\ell}} 
\newcommand{\bM}{\mathbf{M}} 
\newcommand{\bp}{\mathbf{p}} 
\newcommand{\bQ}{\mathbf{Q}} 
\newcommand{\btQ}{\mathbf{\widetilde Q}} 
\newcommand{\bq}{\mathbf{q}} 
\newcommand{\dq}{\dot q} 
\newcommand{\dbq}{\dot{\mathbf{q}}} 
\newcommand{\sq}{\mathsf{q}} 
\newcommand{\dsq}{\dot{\mathsf{q}}} 
\newcommand{\bR}{\mathbf{R}} 
\newcommand{\btR}{\mathbf{\widetilde R}} 
\newcommand{\br}{\mathbf{r}} 
\newcommand{\bS}{\mathbf{S}} 
\newcommand{\bolds}{\mathbf{s}} 
\newcommand{\bT}{\mathbf{T}} 
\newcommand{\bU}{\mathbf{U}} 
\newcommand{\bu}{\mathbf{u}} 
\newcommand{\btu}{\mathbf{\widetilde u}} 
\newcommand{\bv}{\mathbf{v}} 
\newcommand{\bW}{\mathbf{W}} 
\newcommand{\bw}{\mathbf{w}} 
\newcommand{\bX}{\mathbf{X}} 
\newcommand{\bx}{\mathbf{x}} 
\newcommand{\dbx}{\dot{\mathbf{x}}} 
\newcommand{\bY}{\mathbf{Y}} 
\newcommand{\by}{\mathbf{y}} 
\newcommand{\bZ}{\mathbf{Z}} 
\newcommand{\bz}{\mathbf{z}} 
\newcommand{\bkappa}{\bm{\kappa}} 
\newcommand{\bKappa}{\bm{\varkappa}} 
\newcommand{\brho}{\bm{\rho}} 
\newcommand{\bomega}{\bm{\omega}} 
\newcommand{\balpha}{\bm{\alpha}} 
\newcommand{\bxi}{\bm{\xi}} 
\newcommand{\bsigma}{\bm{\sigma}} 
\newcommand{\dvareps}{\dot{\varepsilon}} 
\newcommand{\boldeta}{\bm{\eta}} 

\newcommand{\defeq}{\vcentcolon=} 
\newcommand{\cablecellcolorbot}{red!15}
\newcommand{\barcellcolorbot}{cyan!15}
\newcommand{\cablecellcolortop}{red!30}
\newcommand{\barcellcolortop}{cyan!40}
\newcommand{\cct}{\cellcolor{\cablecellcolortop}}
\newcommand{\bct}{\cellcolor{\barcellcolortop}}


\maketitle
\thispagestyle{empty}
\pagestyle{empty}

\begin{abstract}

Robots built from cable-driven tensegrity (`tension-integrity') structures have many of the advantages of soft robots, such as flexibility and robustness, while still obeying simple statics and dynamics models.
However, existing tensegrity modeling approaches cannot natively describe robots with arbitrary rigid bodies in their tension network.
This work presents a method to calculate the cable tensions in static equilibrium for such tensegrity robots, here defined as compound tensegrity.
First, a static equilibrium model for compound tensegrity robots is reformulated from the standard force density method used with other tensegrity structures.
Next, we pose the problem of calculating tension forces in the robot's cables under our proposed model.
A solution is proposed as a quadratic optimization problem with practical constraints.
Simulations illustrate how this inverse statics optimization problem can be used for both the design and control of two different compound tensegrity applications: a spine robot and a quadruped robot built from that spine.
Finally, we verify the accuracy of the inverse statics model through a hardware experiment, demonstrating the feasibility of low-error open-loop control using our proposed methodology.

\end{abstract}

\section{INTRODUCTION}

Soft robots have been shown to outperform traditional rigid robots in a number of tasks relating to manipulation, human interaction, and locomotion \cite{Majidi2014}.
Rigid robots typically require large, heavy mechanisms and precise actuation alongside sophisticated algorithms for environmental interaction.
In contrast, the natural compliance of soft robots allows for lower mass, bio-inspired actuation, and inherently safe interactions without active control \cite{Laschi2016}.

A recent alternative to either soft or rigid robots are tensegrity (`tension-integrity') systems, built with rigid elements suspended in a network of flexible cables \cite{skelton2009tensegrity}.
Tensegrity robots are compliant \cite{Vespignani2018a} and lightweight \cite{Lessard2016} like soft robots, and can exhibit complex motions with minimal actuation \cite{Sabelhaus2018c}.
In addition, tensegrities can also be modeled with rigid-body statics and dynamics \cite{Juan2008,Tur2009,skelton2009tensegrity}, avoiding complex approaches from continuum mechanics \cite{Giorelli2015}, beam theory \cite{Renda2012a,Renda2012}, or other elastica \cite{DePayrebrune2017a,Goldberg2019} found in soft systems.


\begin{figure}[thpb]
    \centering
    \includegraphics[width=0.9\columnwidth]{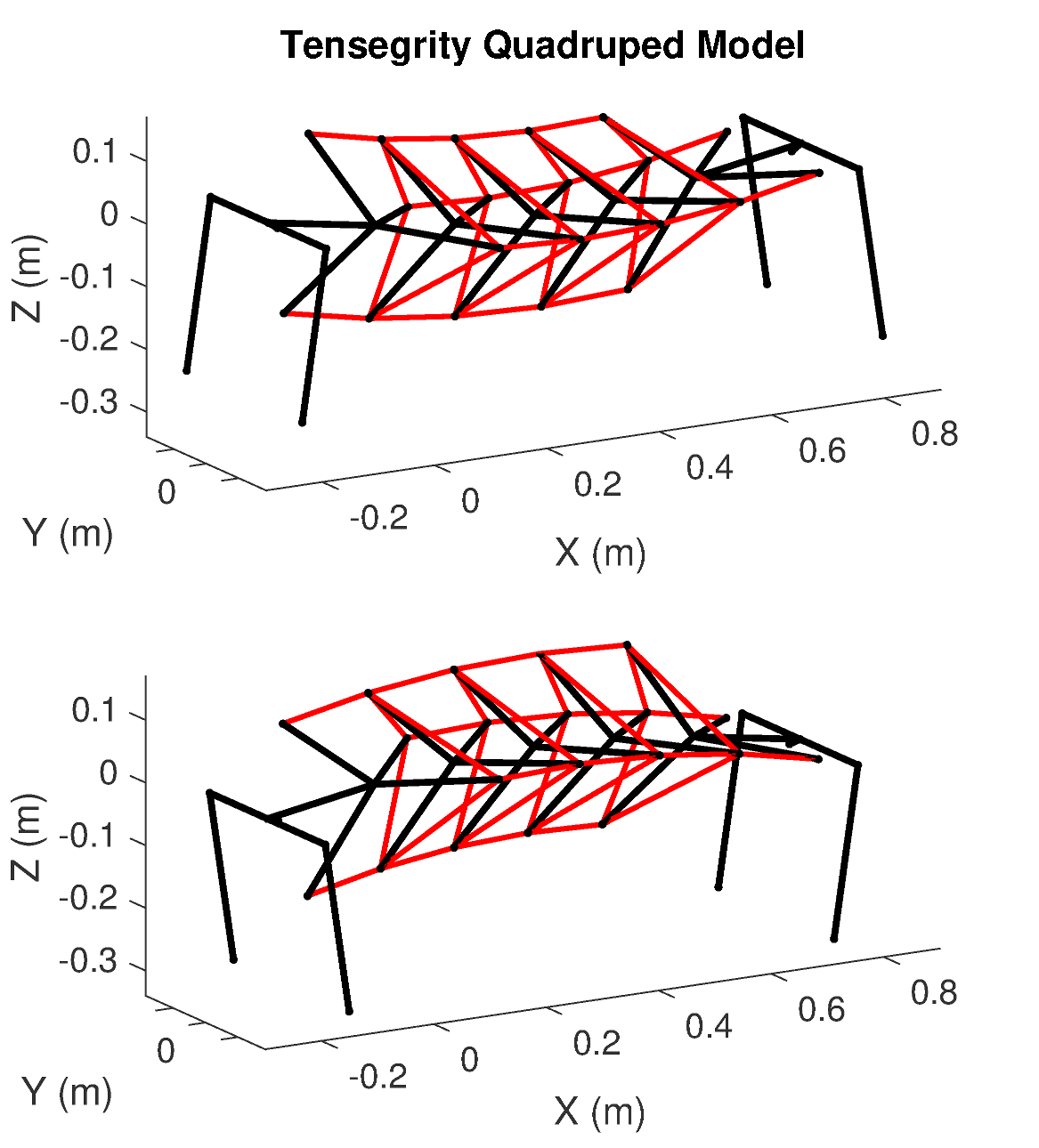}
    \caption{Simulations of a quadruped robot with a tensegrity spine (a compound tensegrity) in extension and flexion poses. The inverse statics optimization program was used to calculate the tensions of the cables (red) that keep the spine vertebrae and hips/shoulders (black) in static equilibrium.}
    \label{fig:belka_overview}
    \vspace{-0.4cm}
\end{figure}

However, current mathematical frameworks for tensegrity robots are limited to single, uniaxial compression members (`bars') in their network \cite{Juan2008,Tur2009,skelton2009tensegrity,Cheong2015}.
Though these `bars-only' robots have shown innovative motions \cite{Bliss2012,Paul2006a,Caluwaerts2014,Rieffel2009}, practical challenges abound.
These designs often result in a large number of bars and cables \cite{Rieffel2009}, making design and fabrication challenging, with potential for much manufacturing error \cite{Chen2017} and extremely tight space constraints on actuators and sensors \cite{Vespignani2018a}.
As a result, many of the more complex `bars-only' tensegrities have only been tested in simulation or artificial testbeds \cite{Bliss2012}.
Most importantly, this paradigm disallows many useful bio-inspired geometries, enforcing artificial constraints on the structures' shapes.

As a result, much recent work has dispensed with this limitation, and instead has used arbitrary rigid bodies in their tension network for closer bio-inspiration \cite{Mirletz2015,Mirletz2015a,Lessard2016,Friesen2018,Chen2019} and simpler actuation \cite{Bohm2016a,Sabelhaus2018c}.
We define these as \textit{compound tensegrities}.
A compound tensegrity can be modeled by bars that are rigidly \textit{compounded} together to create more complex shapes.
To our knowledge, no unifying mathematical model exists for compound tensegrity robots.



This work presents the first results on general physics-based modeling, design, and control systems for compound tensegrity robots.
The contributions presented here include a static equilibrium model (Sec. \ref{sec:static_equilibrium}), equilibrium cable tension optimization process (Sec. \ref{sec:invstat_opt}), simulations for design (Fig. \ref{fig:belka_overview} and Sec. \ref{sec:simulations}), and experimental validation of an open-loop controller on hardware (Sec. \ref{sec:hardware_validation}).
This work is inspired partially by the inverse statics concept presented in \cite{Sabelhaus2019}, and provides a rigorous mathematical basis for the ongoing expansion of the field.

\section{PRIOR WORK}



Tensegrity structures consist of rigid bodies suspended in a network of cables in tension such that the bodies do not contact each other \cite{skelton2009tensegrity}.
For a `Class-1' tensgrity, only disconnected thin bars are allowed as bodies.
Under an expanded definition of `Class-$K$' tensegrities, up to $K$ thin bars can connect at pin joints \cite{Cheong2015}.
Our paper considers the case when a Class-$K$ tensegrity instead has rigid joints where its bars connect, allowing arbitrary compound rigid bodies to exist in the network.
For this reason, compound tensegrities sacrifice the high strength-to-weight ratio found in Class-1-to-$K$ structures, though the forces on most tensegrity robots are sufficiently small to neglect this consideration.

A number of modeling techniques exist for Class-1 and Class-$K$ tensegrities, both for statics \cite{Juan2008} and dynamics \cite{Tur2009}.
The statics models allow designers to simultaneously solve for an equilibrium pose of a tensegrity robot alongside a set of equilibrium cable tensions, a process known as \textit{form-finding} \cite{Tibert2003,Tran2010}.
Solving for the cable tensions given a pose is termed the \textit{inverse statics} problem \cite{Arsenault2006a} and is often solved using the force-density method \cite{Schek1974}, which relies upon the pin-jointed property of Class-$K$ tensegrities absent from compound tensegrities.


This work proposes a static equilibrium model as an initial framework for the design and control of compound tensegrity robots.
Although it has been shown that closed-loop control of tensegrity robots using dynamics models is feasible \cite{Aldrich2003,Wroldsen2009}, statics models have also been successful in open-loop \cite{friesen2014,Kim2015}.
Statics-based control is also used with other soft continuum robots \cite{Giorelli2015}.
This work exploits the structure of compound tensegrities to formulate a quadratic program for the purpose of similar open-loop control.
Though it is possible to transform a compound tensegrity into an equivalent Class-$K$ tensegrity for modeling \cite{Cheong2015}, the proposed method does not require changing the geometry of the robot.




Two different compound tensegrity robots are studied in this work: a spine robot (Fig. \ref{fig:horizontal_1vert_labelled}, \ref{fig:horizSpine_sim}), and a quadruped robot built with that spine (Fig. \ref{fig:belka_overview}).
These robots are under development for assisting with various types of walking locomotion \cite{Sabelhaus2017,Sabelhaus2018c}. 
The inverse statics optimization method presented here is used for control of the spine via bending, and for conducting a design study of pretensioned elastic lattices in the quadruped's body.

\section{STATIC EQUILIBRIUM FOR \\TENSEGRITY ROBOTS}\label{sec:static_equilibrium}

The following section formulates a static equilibrium condition for tensegrity robots such that their cables counteract forces and moments on their rigid bodies.
The force density method is reviewed and then is reformulated for compound tensegrities.
We derive the three-dimensional case here; Sec. \ref{sec:reformulation} notes minor modifications for two dimensions.



\subsection{Definition of a Tensegrity Robot}\label{sec:defn}

Tensegrities can be described by a graph with $n$ nodes and $m$ edges, ordered such that the first $s$ edges represent tensile cables, the last $r$ edges represent thin rigid bars, and $r+s=m$. The graph structure is described by a \textit{connectivity matrix} $\bC \in \mathbb{R}^{(m \times n)}$, where member $i$ connects nodes $j$ and $k$, $j<k$:

\begin{equation}
	\mathbf{C}(a,b) = 
	\begin{cases}
	1 & \text{if} \quad  a=i, b=j \\
	-1 & \text{if} \quad  a=i, b=k \\
	0 & \text{else.}
	\end{cases}
\end{equation}

\begin{figure}[hbtp]
    \centering
    \includegraphics[width=0.9\columnwidth]{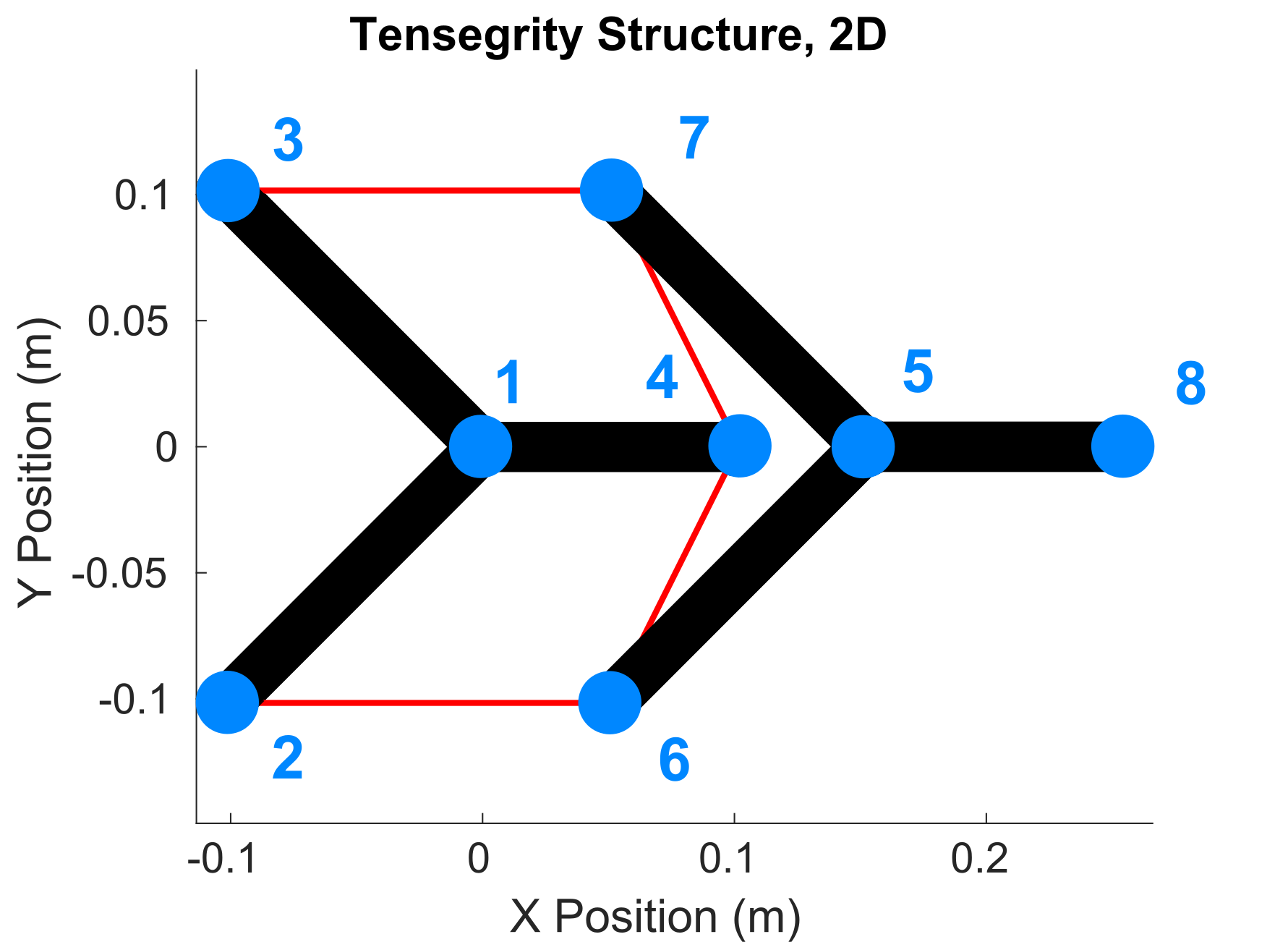}
    \caption{Example two-dimensional, two-vertebra tensegrity spine. Blue circles are nodes, black edges are slender bars (compression), and red edges are cables (tension). }
    \label{fig:horizontal_1vert_labelled}
    \vspace{-0.3cm}
\end{figure}

Each node has a corresponding mapping in $\mathbb{E}^3$ (or $\mathbb{E}^2$ for planar robots).
For example, Fig. \ref{fig:horizontal_1vert_labelled} is a planar projection of a tensegrity spine, with a connectivity matrix of 

\vspace{-0.2cm}
\begin{equation}\label{eqn:connectivity}
\bC
=
\left[ \begin{array}{cccccccc}
  \rowcolor{\cablecellcolorbot} 0 &  1 &  0 &  0 & 0 & -1 &  0 & 0 \\
  \rowcolor{\cablecellcolorbot} 0 &  0 &  1 &  0 & 0 &  0 & -1 & 0 \\
  \rowcolor{\cablecellcolorbot} 0 &  0 &  0 &  1 & 0 & -1 &  0 & 0 \\
  \rowcolor{\cablecellcolorbot} 0 &  0 &  0 &  1 & 0 &  0 & -1 & 0 \\
  \rowcolor{\barcellcolorbot} 1 & -1 &  0 &  0 &  0 &  0 &  0 &  0 \\
  \rowcolor{\barcellcolorbot} 1 &  0 & -1 &  0 &  0 &  0 &  0 &  0 \\
  \rowcolor{\barcellcolorbot} 1 &  0 &  0 & -1 &  0 &  0 &  0 &  0 \\
  \rowcolor{\barcellcolorbot} 0 &  0 &  0 &  0 &  1 & -1 &  0 &  0 \\
  \rowcolor{\barcellcolorbot} 0 &  0 &  0 &  0 &  1 &  0 & -1 &  0 \\
  \rowcolor{\barcellcolorbot} 0 &  0 &  0 &  0 &  1 &  0 &  0 & -1
\end{array} \right],
\end{equation}


\noindent where the red rows represent the $s=4$ cables and blue rows are the $r=6$ bars within the vertebrae.

\subsection{The Force Density Method for Tensegrity Structures}

The force density method (FDM) \cite{Schek1974,Tran2010} is a common approach for calculating equilibrium forces in network structures.
FDM makes the assumptions that (1) all nodes are pin-jointed, (2) external forces are applied only at the nodes, and (3) members are thin and massless. In this paper and others, the mass of a tensegrity robot is modeled as gravitational forces allocated among nodes in the network \cite{friesen2014}.


Given a point for each node in a Cartesian coordinate system $\{x_j,y_j,z_j \}$, the vectors $\bx, \by, \bz\in\mathbb{R}^n$ represent the coordinates for all nodes along the respective axis. 
Similarly, $\bp_x$, $\bp_y$, $\bp_z\in\mathbb{R}^n$ represent the external forces on each node along the respective axis, and $\bp = \begin{bmatrix}\bp_x^\top & \bp_y^\top & \bp_z^\top \end{bmatrix}^\top \in \mathbb{R}^{3n}$. 
Denoting the force in member $i$ as $F_i$, its length as $\ell_i$, and the structure's \textit{force density} vector as

\vspace{-0.3cm}
\begin{equation}
q_i = F_i/\ell_i, \quad \quad \bq = [q_1, \; \hdots, \; q_m]^\top \in \mathbb{R}^m,
\end{equation}
\noindent the equations describing the force balance at all nodes in a network may then be written as in \cite{Schek1974,Tran2010,friesen2014}:

\begin{equation}\label{eqn:init_force_balance}
    \begin{aligned}
    & \bC^\top \text{diag}(\bq) \bC \bx = \bp_x \\
    & \bC^\top \text{diag}(\bq) \bC \by = \bp_y \\
    & \bC^\top \text{diag}(\bq) \bC \bz = \bp_z.
    \end{aligned}
\end{equation}

\noindent Since for two vectors $\mathbf{a},\mathbf{b}$, $\text{diag}(\mathbf{a})\mathbf{b} = \text{diag}(\mathbf{b})\mathbf{a}$, eqn. (\ref{eqn:init_force_balance}) can be reorganized into

\vspace{-0.3cm}
\begin{equation}\label{eqn:Aqp_nodal}
    \bA\bq = \bp,
\end{equation}
\vspace{-0.5cm}

\noindent where

\vspace{-0.1cm}
\begin{equation}\label{eqn:A}
\bA = \begin{bmatrix}
             \bC^\top \text{diag}(\bC\bx) \\
             \bC^\top \text{diag}(\bC\by) \\
             \bC^\top \text{diag}(\bC\bz)
        \end{bmatrix} \in \mathbb{R}^{3n\times m}.
\end{equation}

A value for $\bq$ that meets the constraint of eqn. (\ref{eqn:Aqp_nodal}) may be computed as desired, e.g., with the constrained optimization problem proposed in \cite{friesen2014}, though solutions may not exist in general.



\subsection{Compound Tensegrity Reformulation of the FDM}\label{sec:reformulation}



Examining Fig. \ref{fig:horizontal_1vert_labelled}, moments must be present at nodes 1 and 5, violating assumption (1) for FDM.
For the two robots studied here, eqn. (\ref{eqn:Aqp_nodal}) yields a set of inconsistent equations.


Define instead the compound tensegrity model, where arbitrary moments can evolve at joints where bars rigidly connect to other bars, forming a single rigid body.
A per-body force and moment balance are derived here from eqn. (\ref{eqn:Aqp_nodal}), neglecting the now-internal forces in the bars.

\subsubsection{Force Balance Per-Body}

Assume that $\bC$ is organized according to blocks of connected bars, with $b$ bodies (cf. eqn. (\ref{eqn:connectivity}), block of bars in blue with columns 1-4 and 5-8, $b$=2).
If all bodies have the same number of nodes, define the nodes per body as $\eta = n/b \in \mathbb{Z}_+$ (cf. eqn. (\ref{eqn:connectivity}), $\eta$ = 4).
Otherwise let $\eta_k$ be the number of nodes for body $k$, and $\boldeta = [\eta_1,\hdots, \eta_b]^\top \in \mathbb{Z}_+^b$.
Then, a balance of forces per body is given by the following modification to eqn. (\ref{eqn:Aqp_nodal}):

\vspace{-0.2cm}
\begin{equation}\label{eqn:KaqKp}
\bK \bA \bq = \bK \bp,    
\end{equation}

\noindent where for scalar $\eta$, define the \textit{compounding matrix} $\bK$ as

\vspace{-0.2cm}
\begin{equation}\label{eqn:K_scalar_eta}
    \bK = \bI_{3b} \otimes \bOnes_\eta^\top,
\end{equation}

\noindent or with $\boldeta$,

\vspace{-0.3cm}
\begin{equation}\label{eqn:K_vector_eta}
\bK = \bI_3 \otimes \widetilde \bK, \quad \quad  
    \widetilde \bK = \begin{bmatrix}
    \bOnes_{\eta_1}^\top & \bzero & \\
    \bzero & \ddots & \\
    & & \bOnes_{\eta_b}^\top
    \end{bmatrix} \in \mathbb{R}^{b \times n}.
\end{equation}


Because the bars' force densities are now internal to each rigid body, eqn. (\ref{eqn:KaqKp}) must be further modified to remove them from the network.
Examining the structure of $\bC$, the force densities are implicitly organized into blocks for the cables $\bq_s$ and the bars $\bq_r$, so $\bq = [\bq_s^\top \; \; \bq_r^\top]^\top$.
Then define

\begin{equation}\label{eqn:H_defn} 
\bH = \begin{bmatrix}
    \bI_{s} \\
    \bzero_{r \times s}
\end{bmatrix} \in \mathbb{R}^{(s+r) \times s},
\quad \quad
\bH \bq_s = \begin{bmatrix} \bq_s \\ \bzero_r \end{bmatrix},
\end{equation}

\noindent which can be substituted into eqn. (\ref{eqn:KaqKp}):

\vspace{-0.2cm}
\begin{equation}\label{eqn:KAHqs_Kp}
    \bK \bA \bH \bq_s = \bK \bp,
\end{equation}
\vspace{-0.4cm}

\noindent or equivalently,

\vspace{-0.4cm}
\begin{align}\label{eqn:force_balance_3d}
\bA_f \bq_s = \bp_f, & 
\begin{split}
    \bA_f &= \mathbf{KAH} \\
    \bp_f &= \mathbf{Kp}.
\end{split}
\end{align}

\subsubsection{Moment Balance Per-Body}

Similarly, the moments due to both external forces and cable tensions on each node must now be balanced per rigid body. 
Since this balance can occur in an arbitrary frame, choose the origin.
The moment cross products can be calculated using a skew-symmetric matrix for the moment arms, which are now simply the nodal coordinates,


\begin{equation}
    \bB = \begin{bmatrix}
    \mathbf{0} & -\bZ & \bY \\
    \bZ & \mathbf{0} & -\bX \\
    -\bY & \bX & \mathbf{0}
    \end{bmatrix},
\end{equation}
\vspace{-0.2cm}

\noindent where 

\vspace{-0.3cm}
\begin{equation}
    \bX = \text{diag}(\bx), \quad \quad
    \bY = \text{diag}(\by), \quad \quad
    \bX = \text{diag}(\bz).
\end{equation}

\noindent So, pre-multiplying eqn. (\ref{eqn:Aqp_nodal}) by $\bB$ yields moments applied to a body at each node,



\vspace{-0.1cm}
\begin{equation}\label{eqn:BAqBp}
\bB \bA \bq = \bB \bp.
\end{equation}


As with eqn. (\ref{eqn:force_balance_3d}), the moment balance per body requires compounding rows of eqn. (\ref{eqn:BAqBp}) and removing bar members, so is given by




\vspace{-0.3cm}
\begin{align}\label{eqn:moment_balance_3d}
\bA_m \bq_s = \bp_m, &
\begin{split}
    \bA_m &= \mathbf{KBAH} \\
    \bp_m &= \mathbf{KBp}.
\end{split}
\end{align}




\subsubsection{Combined Compound Static Equilibrium Constraint}

The force and moment balances can be consolidated into a single system of linear equations,

\vspace{-0.2cm}
\begin{equation}\label{eqn:final_fm_balance}
    \bA_b\bq_s = \bp_b,
\end{equation}

\noindent where

\vspace{-0.1cm}
\begin{equation}\label{eqn:final_fm_balance_details}
    \bA_b = \begin{bmatrix}
        \bA_f \\
        \bA_m
    \end{bmatrix} \\, \quad
    \bp_b = \begin{bmatrix}
        \bp_f \\
        \bp_m
    \end{bmatrix}.
\end{equation}

\noindent Eqn. (\ref{eqn:final_fm_balance}) is the static equilibrium constraint used in the remainder of this work, and allows the formulation of a compound tensegrity network as a set of cable-balanced rigid bodies. 
This procedure in the 2D case would instead use $\bB = [-\bY \; \; \bX]$, and $\widetilde \bK$ in place of $\bK$ in eqn. (\ref{eqn:moment_balance_3d}).






\section{INVERSE STATICS OPTIMIZATION}\label{sec:invstat_opt}

Using eqn. (\ref{eqn:final_fm_balance}), an optimal set of cable tensions can be found.
This section presents approaches for finding the nodal applied forces $\bp$ given different designs, a set of constraints for feasibility of cable force densities $\bq_s$ on a physical robot, and a quadratic program to find an optimal $\bq_s^*$.


\subsection{External Forces}\label{sec:ext_rxn_forces}

The external applied forces $\bp$, not due to the cable network, must be specified by the designer.
If not calculated by hand, there are at least two algorithmic ways to find $\bp$ for the given static equilibrium assumptions.

\subsubsection{Removal of Anchor Nodes}\label{sec:removal_anchor_nodes}

First, any \textit{anchor} nodes can be removed from the force and moment balance in eqn. (\ref{eqn:final_fm_balance}), eliminating the need to pre-solve for corresponding entries in $\bp$. 
Doing so allows arbitrary reaction forces to evolve at these nodes, which may be useful for robots in which the tensegrity network interfaces with another part of the system, such as a spine to a hip.
Define an indicator vector $\bw$ whose $i^{th}$ component is 0 if node $i$ is an anchor node:

\vspace{-0.5cm}
\begin{equation}
    \bw \in \{0, 1\}^n, \quad \; \; \bW = \text{nonzero rows}(\text{diag}({\bw})).
\end{equation}


Then, pattern $\bW$ in each direction, as $\bW_f = \bI_3 \otimes \bW$.
Finally, applying $\bW_f$ before compounding per-body removes anchor nodes' forces from eqns. (\ref{eqn:force_balance_3d}) and (\ref{eqn:moment_balance_3d}),

\vspace{-0.2cm}
\begin{align}\label{eqn:balance_anchored_3d}
\begin{split}
    \bA_f &= \bK \bW_f \bA \bH \\
    \bp_f &= \bK \bW_f \bp
\end{split}
&
\begin{split}
    \bA_m &= \bK \bW_f \bB \bA \bH \\
    \bp_m &= \bK \bW_f \bB \bp.
\end{split}
\end{align}



\subsubsection{Pre-Solving at Pinned Joints}\label{sec:presolve_pinned}

Alternatively, solutions for a set of desired reaction forces at nodes $\br \in \mathbb{R}^{nd}$ can be used as part of whole-body motion planning. 
Once $\br$ is calculated, it can be inserted into $\bp$ for the corresponding nodes.
By combining desired reaction forces paired with a pose, the tension network can be constrained to produce, for example, specific frictional interactions.
This approach resembles a quasi-static version of methods such as \cite{Ott2011} for posture control and motion planning.

To calculate a $\br$, let a set of other environmental forces $\bp_{ext}$ at each node be specified (for example, gravitational forces).
Use $\bw$ as before, where where $w(i)$ = 1 indicates a pinned node, to isolate the coordinates of those nodes into $\bx_v = \bW_f \bx,$  similarly for $\by_v$ and $\bz_v$.
Subscripts on the matrices below will then indicate construction with either the full node vectors $\bx,\by,\bz$ (size $n$) or the pinned node vectors $\bx_v,\by_v,\bz_v$ (size $v$) respectively.
The static equilibrium constraint for the reaction forces can be formulated by the same arguments as eqn. (\ref{eqn:final_fm_balance}), specifically

\vspace{-0.2cm}
\begin{equation}\label{eqn:presolve_pinned_3d}
    \bG \br = \bb, \quad \quad 
    \bG = \begin{bmatrix} \bG_f \\ \bG_m \end{bmatrix}, \quad
    \bb = \begin{bmatrix} \bb_f \\ \bb_m \end{bmatrix},
\end{equation}
\vspace{-0.4cm}

\noindent where

\vspace{-0.3cm}
\begin{align}\label{eqn:balance_pinned_3d}
\begin{split}
    \bG_f &= \bK_v \\
    \bb_f &= \bK_n \bp_{ext}
\end{split}
&
\begin{split}
    \bG_m &= \bK_v \bB_v \\
    \bb_m &= \bK_n \bB_n \bp_{ext}.
\end{split}
\end{align}

\noindent Here, indices in $\br$ do not correspond to indices in $\bp$, and so must be tracked and re-inserted separately (example given in accompanying software).





\subsection{Constraints and Objective Function}

With eqn. (\ref{eqn:final_fm_balance}) and an appropriately-calculated $\bp$, an optimization problem can be posed to find the optimal cable tensions satisfying static equilibrium. 
Here, assume a linear elastic model for the system's cables, where cable $i$ has length $\ell_i$, transmits scalar force $F_i$, has linear spring constant $\kappa_i$, and is subject to a controller with input $u_i$ controlling the contraction of the cable:

\vspace{-0.2cm}
\begin{equation}\label{eqn:F_with_input}
    F_i = \kappa_i(\ell_i-u_i).
\end{equation}
\vspace{-0.4cm}





A feasible input $u_i$ requires two main constraints: a minimum tension (equivalently, force density) to avoid slack cables, and a saturation constraint preventing negative cable rest lengths $u_i$.
These constraints provide upper and lower bounds on the optimal force density. 
Define a scalar $c>0$ to be the minimum permissible force density for any cable, and $\bc=c\mathbf{1}_s$. 
The minimum tension constraint is then

\vspace{-0.2cm}
\begin{equation}
    -\bq_s \leq - \bc.
\end{equation}
\vspace{-0.2cm}

Next, to prevent negative rest lengths, it is required that 

\vspace{-0.1cm}
\begin{equation}\label{eqn:u_greater_0}
    u_i > 0, \quad \quad \forall \; i = 1, \dots, s
\end{equation}

\noindent and in practice, the designer should choose a value $u_i^{min}$, or $\mathbf{u}^{min}\in\mathbb{R}^s$ in vector form, for robustness and to enforce physical constraints. 
Substituting eqn. (\ref{eqn:F_with_input}) into (\ref{eqn:u_greater_0}),

\begin{equation}
    u_i^{min} \leq \ell_i - \frac{F_i}{\kappa_i},
\end{equation}

\noindent which can be equivalently written in the linear form

\vspace{-0.1cm}
\begin{equation}\label{eqn:sat_constraint}
    \ell_i q_i \leq \kappa_i(\ell_i - u_i^{min}),
\end{equation}


\noindent or as a system of equations with $\bKappa = \text{diag}(\bkappa)$ and $\bL = \text{diag}(\bell)$,

\vspace{-0.3cm}
\begin{align}
    & \bL \bq_s \leq \bKappa(\bell - \bu^{min}),
\end{align}

\noindent and both constraints can be written compactly as

\begin{equation}\label{eqn:saturation_constraints}
    \begin{bmatrix}
        \bL \\
        -\bI_s
    \end{bmatrix}
    \bq_s \leq 
    \begin{bmatrix}
        \bKappa(\bell - \bu^{min}) \\
        -\bc
    \end{bmatrix}.
\end{equation}

\noindent 



In addition, the total potential energy in the system can be formulated via eqn. (\ref{eqn:F_with_input}). 
Using this as an objective function minimizes energy requirements for control.
As a function of the control input, the potential energy in the cable system is

\vspace{-0.2cm}
\begin{equation}\label{eqn:linear_elastic_potential_energy}
PE = \frac{1}{2} \sum_{i=1}^s \kappa_i (\ell_i - u_i)^2.
\end{equation}

\noindent Substituting from eqn. (\ref{eqn:F_with_input}) and reorganizing,

\begin{equation}
    PE_i = \frac{1}{2} q_i^2 \frac{\ell_i^2}{\kappa_i}.
\end{equation}

\noindent Therefore, the total potential energy is quadratic in $\bq_s$, as

\begin{equation}\label{eqn:linear_elastic_total_PE}
    PE = \frac{1}{2} \bq_s^\top \bKappa^{-1} \bL^2 \bq_s.
\end{equation}

\subsection{Solution via a Quadratic Program}

With the above quadratic cost and linear constraints, the following quadratic program can be used to solve for the optimal cable tensions for one pose of the robot.
This procedure is defined here as \textit{inverse statics optimization}:

\vspace{-0.3cm}
\begin{align}
    \bq_s^* = \argmin_{\bq_s} \; \; & \bq_s^\top \bR \bq_s \label{eqn:optprob}\\
    \text{s.t.} \; \; & \bA_b \bq_s = \bp_b \\
                      & \bS \bq_s \leq \bv, \label{eqn:Sqv}
\end{align}
\vspace{-0.5cm}

\noindent with

\vspace{-0.3cm}
\begin{equation}\label{eqn:constraints}
    \bR = \bKappa^{-1} \bL^2, \quad
    \bS = \begin{bmatrix} \bL \\ -\bI_s \end{bmatrix}, \quad
    \bv = \begin{bmatrix} \bKappa(\bell - \bu^{min}) \\ -\bc \end{bmatrix}.
\end{equation}


\noindent An implementation of the above is provided by the authors in MATLAB, using the {\tt{quadprog}} solver\footnote{https://github.com/apsabelhaus/tiso}.

Similar optimization problems have been proposed for Class-1-to-K tensegrity robots \cite{friesen2014}.
However, this is the first approach that solves for the equilibrium tensions of compound tensegrities and the first to address both energy and saturation constraints (as in eqn. (\ref{eqn:saturation_constraints})).


For purposes of design, the optimization program (\ref{eqn:optprob}-\ref{eqn:constraints}) can be used to generate an optimal $\bq_s^*$ in a single specified pose of the tensegrity robot, i.e., one position of $\bx,\by,\bz$. 
For control, (\ref{eqn:optprob}-\ref{eqn:constraints}) can instead be used in a series of poses, $\bxi(t)$, $t=1\hdots T$, where $\bxi$ specifies the position and orientation of each rigid body, from which all nodal positions may be determined. 
Open-loop control can therefore be performed by solving for a $\bq_s^*(t)$ and applying the corresponding control inputs $\bu$, as motivated by other authors \cite{friesen2014}.

\section{SIMULATIONS}\label{sec:simulations}




Static equilibrium cable tensions in a tensegrity robot provide a convenient simplification from dynamic behavior for pseudo-static open-loop control \cite{friesen2014}.
In addition, manufacturing a set of elastic cables for a tensegrity robot (such as in \cite{Chen2017}) requires determining an initial pose, implicitly in static equilibrium.
Our proposed quadratic program (\ref{eqn:optprob}-\ref{eqn:constraints}) provides a fast, exact solution to this problem for compound tensegrity robots where previous techniques could not.
This section presents sets of simulations using the proposed solution for these purposes.
First, a version of the tensegrity spine from Fig. \ref{fig:horizontal_1vert_labelled} with multiple vertebra was simulated for open-loop control, and second, the tensegrity quadruped from Fig. \ref{fig:belka_overview} was simulated for mechanical design.
Implementation details for both sets of simulations, such as the trajectories of poses $\bxi(t)$, can be found in the accompanying software for this paper$^1$.

\begin{figure}[thpb]
    \centering
    \includegraphics[width=0.95\columnwidth]{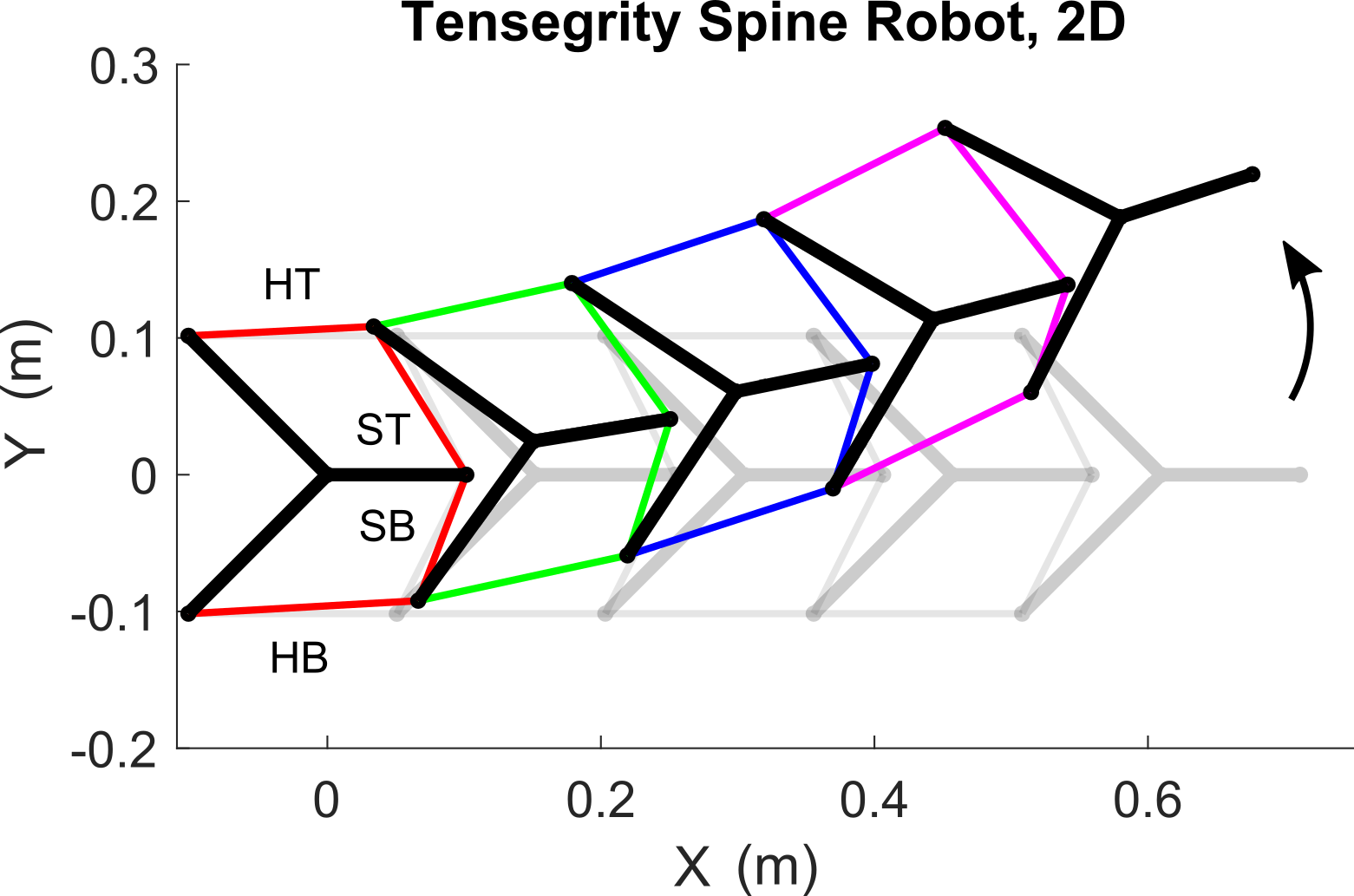}
    \caption{Simulation of a 2D tensegrity spine robot in a bending motion, cables colorized according to set of vertebra, labelled within one set. Initial pose is horizontal (gray).}
    \label{fig:horizSpine_sim}
\end{figure}

\begin{figure}[thpb]
    \centering
    \includegraphics[width=0.95\columnwidth]{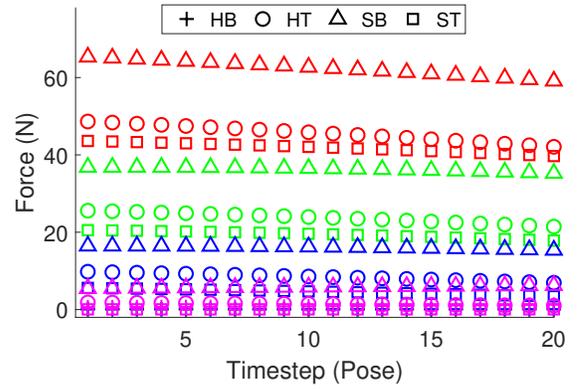}
    \caption{Inverse statics optimization results for the bending 2D spine from Fig. \ref{fig:horizSpine_sim}, same coloring and labeling. Significantly larger trends exist between cables than between poses.}
    \label{fig:horizSpine_sim_data}
    \vspace{-0.3cm}
\end{figure}

\subsection{Two-Dimensional Tensegrity Spine Robot}

The inverse statics optimization problem was first used to generate simulated control inputs for a five-vertebra version of the planar tensegrity spine (Fig. \ref{fig:horizontal_1vert_labelled}) discussed in Sec. \ref{sec:defn}.
The spine performed an upward bending motion, which we have previously used to show beneficial movements of a quadruped's body and feet \cite{Sabelhaus2017,Sabelhaus2018c}.
The leftmost vertebra was considered anchored, and the approach from Sec. \ref{sec:removal_anchor_nodes} was used.
The model used only gravitational forces in $\bp$ due to a mass of 0.495 kg distributed to all nodes, as well as a spring constant of $\kappa_i=4.8$ lbf/in.
Optimization constants were $c=0.5$ N/m, $u_i^{min}=0.01$ m.
As shown in Fig. \ref{fig:horizSpine_sim}, the spine traversed $T = 20$ poses from a horizontal initial configuration to an upward-bent final configuration.



The inverse statics optimization problem generated the sets of optimal forces shown in Fig. \ref{fig:horizSpine_sim_data}.
Cables were organized by color into sets between vertebra from left to right: red, green, blue, magenta.
Cables within a set were named with the ordering from $\bC$ in eqn. (\ref{eqn:connectivity}): ``horizontal'' top/bottom, ``saddle'' bottom/top: HT, HB, SB, ST.


Solutions existed in all poses of the robot.
Since the spine is cantilevered, the leftmost cables required much higher tensions to support the weight of the rest of the structure.
Small variations in tensions caused large changes in pose.
The horizontal-bottom cables were optimally tensioned at $u_i^{min}$, with the horizontal-top and saddle-bottom exerting the most force for this counterclockwise bend.

\begin{figure}[bpht]
    \centering
    \includegraphics[width=0.95\columnwidth]{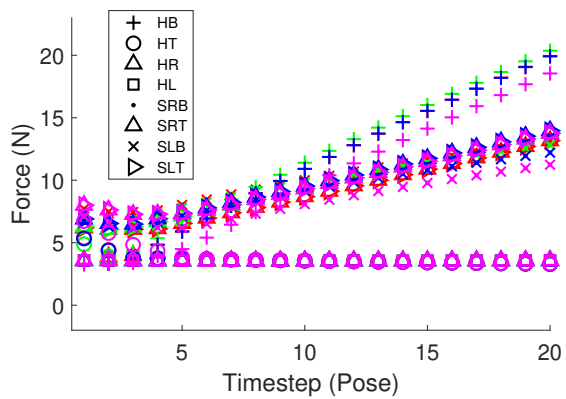}
    \caption{Inverse statics optimization results for the quadruped bending motion from Fig. \ref{fig:belka_overview}. Robot moves from sagittal extension at $t$=0 to sagittal flexion at $t$=20 (Fig. \ref{fig:belka_overview} top vs. bottom). Unlike the 2D spine, tensions change dramatically during the motion, from evenly tensioned when extended to mostly ventral tension (HB) when flexed.}
    \label{fig:belkaBending_sim_data}
    \vspace{-0.4cm}
\end{figure}

\subsection{Three-Dimensional Tensegrity Quadruped}\label{sec:quadruped_sim}

The inverse statics calculations were also used for design of a tensegrity quadruped.
The quadruped robot Laika uses a tensegrity spine as its body \cite{Sabelhaus2018c}, which in turn is constructed using a pretensioned lattice of elastic material as per the techniques in \cite{Chen2017}.
The authors' initial prototypes of Laika from \cite{Sabelhaus2018c} used hand-chosen tensions in the elastic lattice for purposes of rapid prototyping.
This paper introduces a design study of lattice tensions for specific robot poses.

The quadruped robot in Fig. \ref{fig:belka_overview} models an updated version of Laika, with total length of 95 cm and a mass of 7.3 kg.
A series of standing poses were chosen for simulation, with the robot's back arched to varying degrees (Fig. \ref{fig:belka_overview} top and bottom), intended for studying interactions between spine tensioning and locomotion gaits in future work.

The external force vector $\bp$ was calculated by pre-solving for reaction forces at the quadruped's feet, treating them as pinned joints as per Sec. \ref{sec:presolve_pinned}.
Additional constraints were added such that $\bR_x$=0 and $\bR_y$=0, preventing frictional forces from evolving at the feet (see accompanying software).
The optimization used $c = 25$ N/m, without the $\bu^{min}$ constraint.

The optimal cable tensions are shown in Fig. \ref{fig:belkaBending_sim_data}, proceeding from spine extension at $t$=0 (Fig. \ref{fig:belka_overview} top) to flexion at $t=T$ (Fig. \ref{fig:belka_overview} bottom).
Cables were labeled similarly to Fig. \ref{fig:horizSpine_sim_data}: horizontal/saddle top, bottom, left, and right.
Unlike the bending spine, large changes in cable forces were seen as the quadruped is pretensioned in different poses.
Cable forces were distributed somewhat evenly when in extension, but in flexion, the horizontal-bottom cables (along the ventral edge) supported the majority of the robot's mass.

The results from Fig. \ref{fig:belkaBending_sim_data} inform the design of future elastic lattices for Laika.
Generally, higher pretensions produce higher stiffness in the robot.
For walking gaits that require flexibility, a pretensioned cable network with an extended spine in equilibrium may be preferable, whereas if structural stability is required, the flexed spine may be best.

\section{HARDWARE VALIDATION}\label{sec:hardware_validation}





\begin{figure}[bhtp]
    \centering
    \includegraphics[width=1\columnwidth]{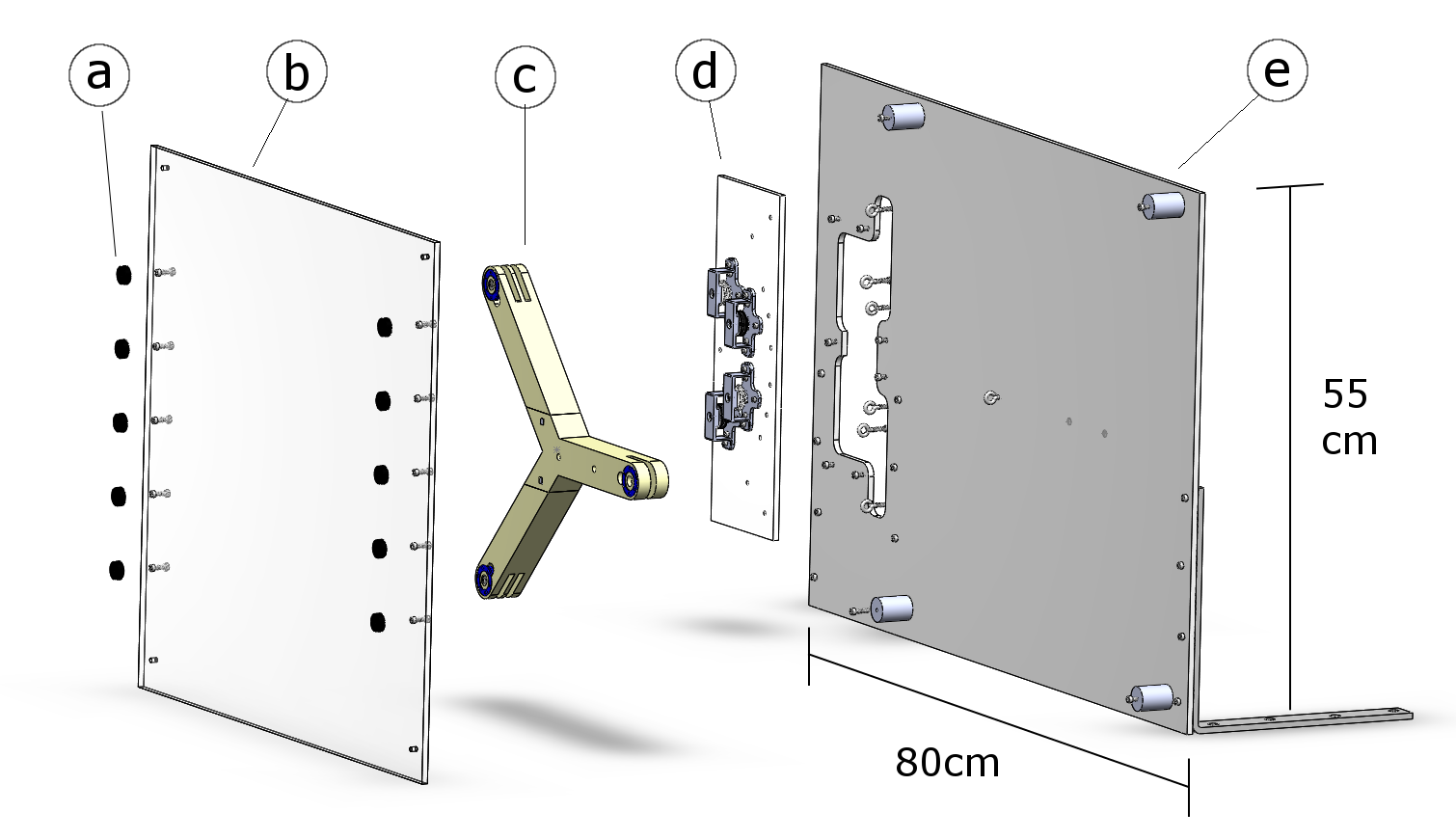}
    \caption{CAD render of the test setup. (a) Calibration markers for the computer vision system, (b) front acrylic plate, (c) vertebra, (d) motor assembly, (e) rear plate.}
    \label{fig:cad_render}
    \vspace{-0.3cm}
\end{figure}

Finally, we conducted a hardware test of pseudo-static open-loop control for the two-dimensional spine, with a single vertebra, to validate the accuracy of the static equilibrium formulation.
The test setup had the same relative geometry as Fig. \ref{fig:horizontal_1vert_labelled}, but was scaled to be larger.
It consisted of a 3D-printed vertebra with $m=$ 0.495 kg, sized approximately 20 cm x 20 cm, constrained between two plastic plates (Fig. \ref{fig:cad_render}).
A camera connected to a computer running OpenCV was pointed at the clear acrylic front plate, and tracked various markers on the vertebra.
A microcontroller controlled the position of four brushless 30W DC motors from Maxon to adjust the lengths of the four cables.



\begin{figure}[bhpt]
    \centering
    \includegraphics[width=0.85\columnwidth]{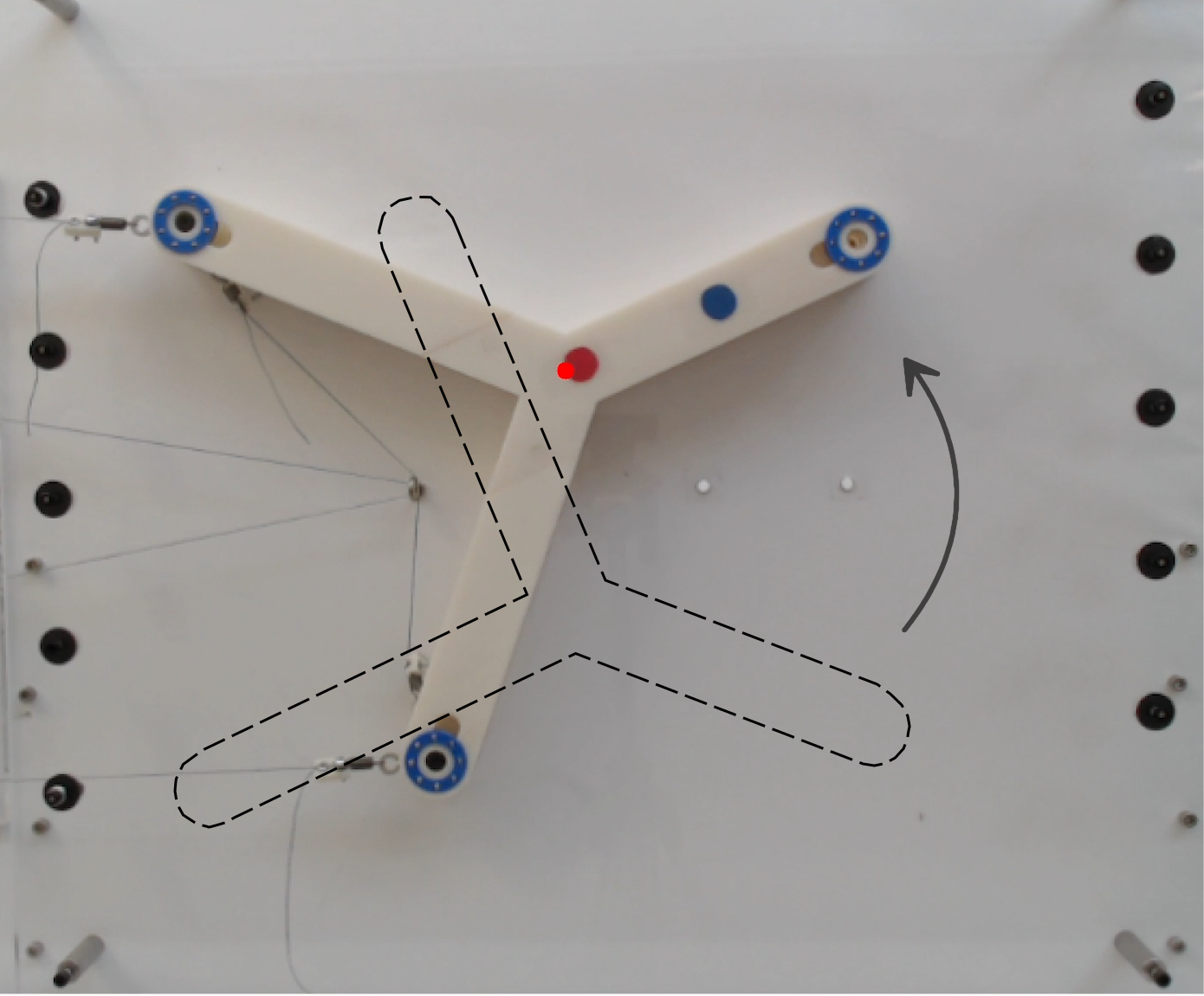}
    \caption{Representative hardware test, final pose (initial pose in dotted outline). Red dot indicates the center of mass of the vertebra.}
    \label{fig:hw_test_combined}
    \vspace{-0.4cm}
\end{figure}

The test procedure consisted of first pinning the vertebra into a pre-specified location, then tensioning each cable by hand to approximately zero force.
This calibration was repeated between each test.
Then, over the course of 80 seconds, control inputs from the inverse statics optimization program were sent to the microcontroller, resulting in a bending motion of the spine (Fig. \ref{fig:hw_test_combined}).
Ten repeated tests were performed, and for each test, the position and rotation of the vertebra were post-processed from the computer vision data.

\begin{figure}[thpb]
    \centering
    \includegraphics[width=1.05\columnwidth]{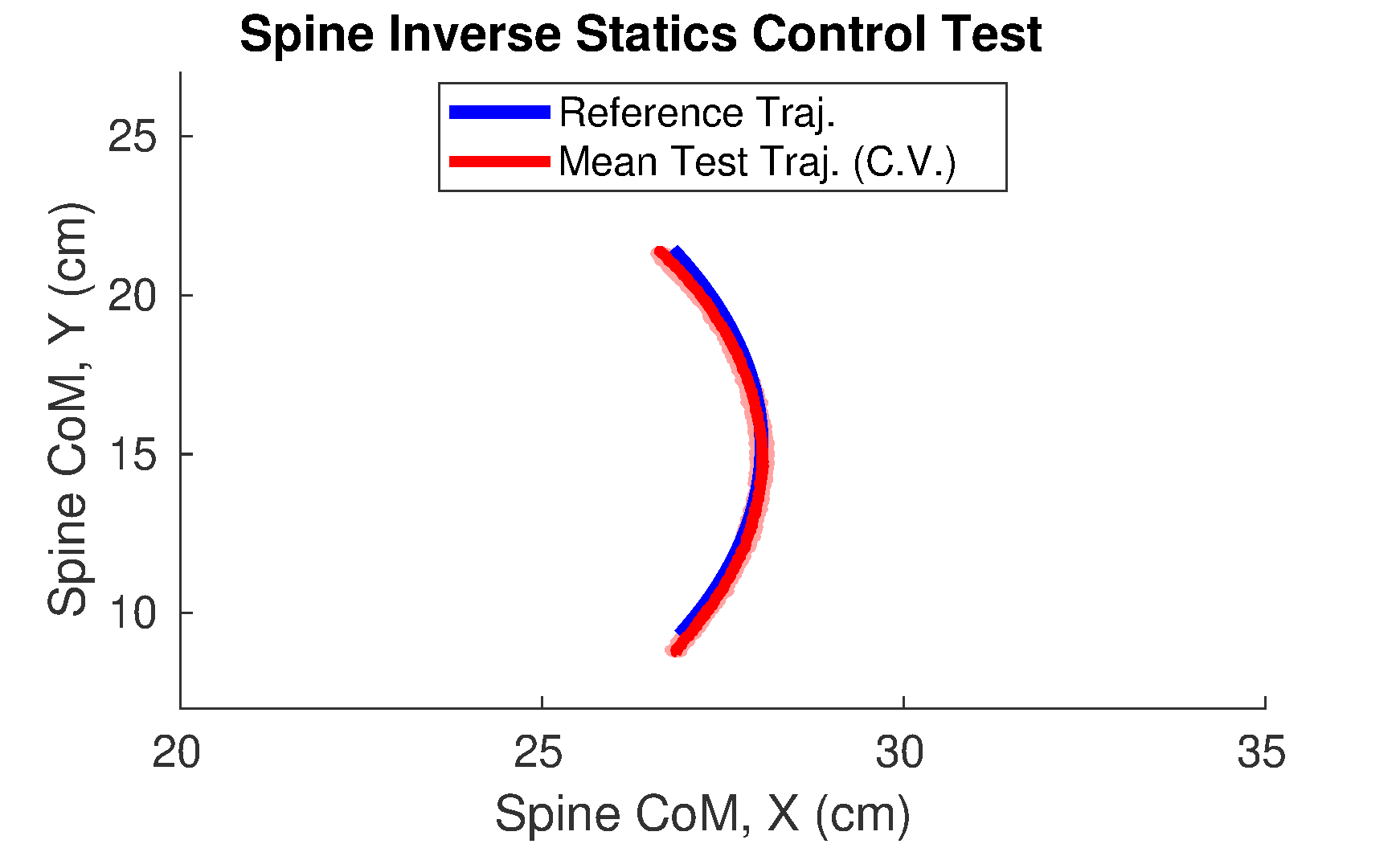}
    \caption{Averaged trajectory of the center of mass of the vertebra, over 10 trials. The spine moved along its intended path with extremely small error.}
    \label{fig:com_traj}
    \vspace{-0.3cm}
\end{figure}

\begin{figure}[thpb]
    \centering
    \includegraphics[width=0.9\columnwidth]{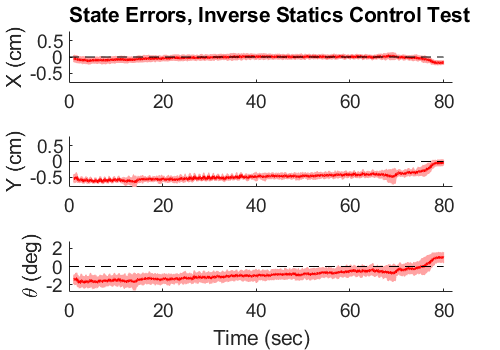}
    \caption{Tracking errors of the vertebra's states, over 10 trials. Red is mean, pink bounds are one standard deviation. Errors are on the scale of the resolution of the camera (1 mm).}
    \label{fig:state_err}
    \vspace{-0.1cm}
\end{figure}

Fig. \ref{fig:com_traj} and \ref{fig:state_err} show the position of the vertebra's center of mass, and the error between the intended trajectory and the measured result.
The mean result over the 10 trials is shown in red, and the pink shaded regions show one standard deviation.
Errors were extremely small, three orders-of-magnitude less than the size of the test setup and length of the trajectory.
In addition, the state errors of the center of mass position were at or around the camera resolution of approximately 1 mm.

\section{DISCUSSION AND CONCLUSION}

The open-loop control test from Sec. \ref{sec:hardware_validation} demonstrates that the proposed inverse statics optimization problem can generate physically accurate cable tensions to hold a compound tensegrity robot in equilibrium poses.
The few sources of error in the test, including friction and calibration, caused only very small drift in the vertebra's path, as is expected in open loop.
Therefore, it is expected that the 3D inverse statics program would perform similarly well (e.g., for a manufactured prototype lattice of a tensegrity quadruped).




To the best of our knowledge, these results provide the first general mathematical model of compound tensegrity robots.
This model is supported by simulations showing its use for control and design, in addition to a hardware experiment validating these calculations.
The approach opens the field to novel designs and control systems for compound tensegrity robots that are grounded by a unifying theoretical model, and by effect, expands the possibilities for innovative solutions to pressing problems in soft robotics.

Though the proposed inverse statics optimization routine is fast (as a quadratic program), easy to formulate (in comparison to approaches such as \cite{Giorelli2015}), and apparently physically valid, it has limitations. 
This statics model does not lend itself to feedback control, and only applies under the specific conditions of compound tensegrity robots.
Models for the standard Class-1-to-$K$ tensegrity robots are much more advanced.

Consequently, future work on compound tensegrity robot modeling will include the structures' dynamics, with a goal to create a general model in the same vein as \cite{skelton2009tensegrity}.
In addition, although statics-based open-loop control has worked well for other tensegrities \cite{friesen2014}, future research will extend to closed-loop control systems for compound tensegrity robots.
Techniques as diverse as \cite{Aldrich2003,Wroldsen2009,Sabelhaus2019} will be considered.
Finally, these models will be used for design studies of locomotion for quadruped robots such as \cite{Sabelhaus2018c}.





\section*{ACKNOWLEDGMENT}


This work would not have been possible without the help of the many members of the Berkeley Emergent Space Tensegrities Lab and the Dynamic Tensegrity Robotics Lab at NASA Ames Research Center's Intelligent Robotics Group.
This research was supported by NASA Space Technology Research Fellowship no. NNX15AQ55H.

\balance

\bibliographystyle{IEEEtran}
\bibliography{bibliographies/library}

\begin{thebibliography}{10}
\providecommand{\url}[1]{#1}
\csname url@samestyle\endcsname
\providecommand{\newblock}{\relax}
\providecommand{\bibinfo}[2]{#2}
\providecommand{\BIBentrySTDinterwordspacing}{\spaceskip=0pt\relax}
\providecommand{\BIBentryALTinterwordstretchfactor}{4}
\providecommand{\BIBentryALTinterwordspacing}{\spaceskip=\fontdimen2\font plus
\BIBentryALTinterwordstretchfactor\fontdimen3\font minus
  \fontdimen4\font\relax}
\providecommand{\BIBforeignlanguage}[2]{{%
\expandafter\ifx\csname l@#1\endcsname\relax
\typeout{** WARNING: IEEEtran.bst: No hyphenation pattern has been}%
\typeout{** loaded for the language `#1'. Using the pattern for}%
\typeout{** the default language instead.}%
\else
\language=\csname l@#1\endcsname
\fi
#2}}
\providecommand{\BIBdecl}{\relax}
\BIBdecl

\bibitem{Majidi2014}
C.~Majidi, ``{Soft Robotics: A Perspective - Current Trends and Prospects for
  the Future},'' \emph{Soft Robotics}, mar 2014.

\bibitem{Laschi2016}
C.~Laschi, B.~Mazzolai, and M.~Cianchetti, ``{Soft robotics: Technologies and
  systems pushing the boundaries of robot abilities},'' \emph{Science
  Robotics}, dec 2016.

\bibitem{skelton2009tensegrity}
R.~E. Skelton and M.~C. de~Oliveira, \emph{{Tensegrity systems}}.\hskip 1em
  plus 0.5em minus 0.4em\relax Springer, 2009.

\bibitem{Vespignani2018a}
M.~Vespignani, J.~M. Friesen, V.~SunSpiral, and J.~Bruce, ``{Design of
  SUPERball v2, a Compliant Tensegrity Robot for Absorbing Large Impacts},'' in
  \emph{2018 IEEE/RSJ International Conference on Intelligent Robots and
  Systems (IROS)}.\hskip 1em plus 0.5em minus 0.4em\relax IEEE, oct 2018.

\bibitem{Lessard2016}
S.~Lessard, J.~Bruce, E.~Jung, M.~Teodorescu, V.~SunSpiral, and A.~Agogino,
  ``{A lightweight, multi-axis compliant tensegrity joint},'' in \emph{2016
  IEEE International Conference on Robotics and Automation (ICRA)}.\hskip 1em
  plus 0.5em minus 0.4em\relax IEEE, may 2016.

\bibitem{Sabelhaus2018c}
A.~P. Sabelhaus, L.~{Janse van Vuuren}, A.~Joshi, E.~Zhu, H.~J. Garnier, K.~A.
  Sover, J.~Navarro, A.~K. Agogino, and A.~M. Agogino, ``{Design, Simulation,
  and Testing of a Flexible Actuated Spine for Quadruped Robots},'' in
  \emph{arXiv:1804.06527}, 2018.

\bibitem{Juan2008}
S.~H. Juan and J.~M.~M. Tur, ``{Tensegrity frameworks: Static analysis
  review},'' \emph{Mechanism and Machine Theory}, 2008.

\bibitem{Tur2009}
J.~M. {Mirats Tur} and S.~H. Juan, ``{Tensegrity frameworks: Dynamic analysis
  review and open problems},'' \emph{Mechanism and Machine Theory}, jan 2009.

\bibitem{Giorelli2015}
M.~Giorelli, F.~Renda, M.~Calisti, A.~Arienti, G.~Ferri, and C.~Laschi,
  ``{Neural Network and Jacobian Method for Solving the Inverse Statics of a
  Cable-Driven Soft Arm with Nonconstant Curvature},'' \emph{IEEE Transactions
  on Robotics}, 2015.

\bibitem{Renda2012a}
F.~Renda and C.~Laschi, ``{A general mechanical model for tendon-driven
  continuum manipulators},'' in \emph{2012 IEEE International Conference on
  Robotics and Automation}.\hskip 1em plus 0.5em minus 0.4em\relax IEEE, may
  2012.

\bibitem{Renda2012}
F.~Renda, M.~Cianchetti, M.~Giorelli, A.~Arienti, and C.~Laschi, ``{A 3D
  steady-state model of a tendon-driven continuum soft manipulator inspired by
  the octopus arm},'' \emph{Bioinspiration {\&} Biomimetics}, jun 2012.

\bibitem{DePayrebrune2017a}
K.~M. de~Payrebrune and O.~M. O'Reilly, ``{On the development of rod-based
  models for pneumatically actuated soft robot arms: A five-parameter
  constitutive relation},'' \emph{International Journal of Solids and
  Structures}, aug 2017.

\bibitem{Goldberg2019}
N.~N. Goldberg, X.~Huang, C.~Majidi, A.~Novelia, O.~M. O'Reilly, D.~A. Paley,
  and W.~L. Scott, ``{On Planar Discrete Elastic Rod Models for the Locomotion
  of Soft Robots},'' \emph{Soft Robotics}, may 2019.

\bibitem{Cheong2015}
J.~Cheong and R.~E. Skelton, ``{Nonminimal Dynamics of General Class k
  Tensegrity Systems},'' \emph{International Journal of Structural Stability
  and Dynamics}, mar 2015.

\bibitem{Bliss2012}
T.~Bliss, J.~Werly, T.~Iwasaki, and H.~Bart-Smith, ``{Experimental Validation
  of Robust Resonance Entrainment for CPG-Controlled Tensegrity Structures},''
  \emph{IEEE Transactions on Control Systems Technology}, may 2013.

\bibitem{Paul2006a}
C.~Paul, F.~Valero-Cuevas, and H.~Lipson, ``{Design and control of tensegrity
  robots for locomotion},'' \emph{IEEE Transactions on Robotics}, oct 2006.

\bibitem{Caluwaerts2014}
K.~Caluwaerts, J.~Despraz, A.~Iscen, A.~P. Sabelhaus, J.~Bruce, B.~Schrauwen,
  and V.~SunSpiral, ``{Design and control of compliant tensegrity robots
  through simulation and hardware validation},'' \emph{Journal of The Royal
  Society Interface}, jul 2014.

\bibitem{Rieffel2009}
J.~A. Rieffel, F.~J. Valero-Cuevas, and H.~Lipson, ``{Morphological
  communication: exploiting coupled dynamics in a complex mechanical structure
  to achieve locomotion},'' \emph{Journal of The Royal Society Interface}, apr
  2010.

\bibitem{Chen2017}
L.-H. Chen, M.~C. Daly, A.~P. Sabelhaus, L.~A. {Janse van Vuuren}, H.~J.
  Garnier, M.~I. Verdugo, E.~Tang, C.~U. Spangenberg, F.~Ghahani, A.~K.
  Agogino, and A.~M. Agogino, ``{Modular Elastic Lattice Platform for Rapid
  Prototyping of Tensegrity Robots},'' in \emph{ASME International Design
  Engineering Technical Conference (IDETC) 41st Mechanisms and Robotics
  Conference}, 2017.

\bibitem{Mirletz2015}
B.~T. Mirletz, I.-W. Park, R.~D. Quinn, and V.~SunSpiral, ``{Towards bridging
  the reality gap between tensegrity simulation and robotic hardware},'' in
  \emph{2015 IEEE/RSJ International Conference on Intelligent Robots and
  Systems (IROS)}.\hskip 1em plus 0.5em minus 0.4em\relax IEEE, sep 2015.

\bibitem{Mirletz2015a}
B.~T. Mirletz, P.~Bhandal, R.~D. Adams, A.~K. Agogino, R.~D. Quinn, and
  V.~SunSpiral, ``{Goal-Directed CPG-Based Control for Tensegrity Spines with
  Many Degrees of Freedom Traversing Irregular Terrain},'' \emph{Soft
  Robotics}, dec 2015.

\bibitem{Friesen2018}
J.~M. Friesen, J.~L. Dean, T.~Bewley, and V.~Sunspiral, ``{A
  Tensegrity-Inspired Compliant 3-DOF Compliant Joint},'' in \emph{2018 IEEE
  International Conference on Robotics and Automation (ICRA)}.\hskip 1em plus
  0.5em minus 0.4em\relax IEEE, may 2018.

\bibitem{Chen2019}
B.~Chen and H.~Jiang, ``{Swimming Performance of a Tensegrity Robotic Fish},''
  \emph{Soft Robotics}, apr 2019.

\bibitem{Bohm2016a}
V.~Bohm, T.~Kaufhold, F.~Schale, and K.~Zimmermann, ``{Spherical mobile robot
  based on a tensegrity structure with curved compressed members},'' in
  \emph{2016 IEEE International Conference on Advanced Intelligent Mechatronics
  (AIM)}.\hskip 1em plus 0.5em minus 0.4em\relax IEEE, jul 2016.

\bibitem{Sabelhaus2019}
A.~P. Sabelhaus, H.~Zhao, E.~Zhu, A.~K. Agogino, and A.~M. Agogino,
  ``{Model-Predictive Control with Inverse Statics Optimization for Tensegrity
  Spine Robots},'' in \emph{arXiv:1806.08868}, 2019.

\bibitem{Tibert2003}
A.~Tibert and S.~Pellegrino, ``{Review of Form-Finding Methods for Tensegrity
  Structures},'' \emph{International Journal of Space Structures}, dec 2003.

\bibitem{Tran2010}
H.~C. Tran and J.~Lee, ``{Advanced form-finding of tensegrity structures},''
  \emph{Computers {\&} Structures}, feb 2010.

\bibitem{Arsenault2006a}
M.~Arsenault and C.~M. Gosselin, ``{Kinematic, Static, and Dynamic Analysis of
  a Planar One-Degree-of-Freedom Tensegrity Mechanism},'' \emph{Journal of
  Mechanical Design}, 2005.

\bibitem{Schek1974}
H.~J. Schek, ``{The force density method for form finding and computation of
  general networks},'' \emph{Computer Methods in Applied Mechanics and
  Engineering}, jan 1974.

\bibitem{Aldrich2003}
J.~Aldrich, R.~Skelton, and K.~Kreutz-Delgado, ``{Control synthesis for a class
  of light and agile robotic tensegrity structures},'' in \emph{Proceedings of
  the 2003 American Control Conference, 2003.}\hskip 1em plus 0.5em minus
  0.4em\relax IEEE, 2003.

\bibitem{Wroldsen2009}
A.~Wroldsen, M.~de~Oliveira, and R.~Skelton, ``{Modelling and control of
  non-minimal non-linear realisations of tensegrity systems},''
  \emph{International Journal of Control}, mar 2009.

\bibitem{friesen2014}
J.~Friesen, A.~Pogue, T.~Bewley, M.~de~Oliveira, R.~Skelton, and V.~Sunspiral,
  ``{DuCTT: A tensegrity robot for exploring duct systems},'' in \emph{2014
  IEEE International Conference on Robotics and Automation (ICRA)}.\hskip 1em
  plus 0.5em minus 0.4em\relax IEEE, may 2014.

\bibitem{Kim2015}
K.~Kim, A.~K. Agogino, A.~Toghyan, D.~Moon, L.~Taneja, and A.~M. Agogino,
  ``{Robust learning of tensegrity robot control for locomotion through
  form-finding},'' in \emph{2015 IEEE/RSJ International Conference on
  Intelligent Robots and Systems (IROS)}.\hskip 1em plus 0.5em minus
  0.4em\relax IEEE, sep 2015.

\bibitem{Sabelhaus2017}
A.~P. Sabelhaus, A.~K. Akella, Z.~A. Ahmad, and V.~SunSpiral,
  ``{Model-Predictive Control of a Flexible Spine Robot},'' in \emph{2017
  American Control Conference (ACC)}.\hskip 1em plus 0.5em minus 0.4em\relax
  IEEE, may 2017.

\bibitem{Ott2011}
C.~Ott, M.~A. Roa, and G.~Hirzinger, ``{Posture and balance control for biped
  robots based on contact force optimization},'' in \emph{2011 11th IEEE-RAS
  International Conference on Humanoid Robots}.\hskip 1em plus 0.5em minus
  0.4em\relax IEEE, oct 2011.

\end{thebibliography}



\end{document}